\documentclass[sigconf]{acmart}
\AtBeginDocument{%
  }

\copyrightyear{2026}
\acmYear{2026}
\setcopyright{cc}
\setcctype{by-nc-nd}
\acmConference[SIGIR '26]{Proceedings of the 49th International ACM SIGIR Conference on Research and Development in Information Retrieval}{July 20--24, 2026}{Melbourne, VIC, Australia}
\acmBooktitle{Proceedings of the 49th International ACM SIGIR Conference on Research and Development in Information Retrieval (SIGIR '26), July 20--24, 2026, Melbourne, VIC, Australia}
\acmDOI{10.1145/3805712.3809696}
\acmISBN{979-8-4007-2599-9/2026/07}

\usepackage{booktabs}   
\usepackage{multirow}   
\usepackage{siunitx}    
\usepackage{makecell}   
\usepackage{pgfplotstable}
\usepackage{amsmath}
\usepackage{xparse}
\usepackage[normalem]{ulem}
\usepackage[x11names, table]{xcolor}
\usepackage{algorithm}
\usepackage{algpseudocode}

\pgfplotsset{compat=1.17}

\ExplSyntaxOn
\NewDocumentCommand{\getrankedvalue}{mm}
 {
  \seq_set_from_clist:Nn \l_tmpa_seq { #2 }
  \seq_remove_duplicates:N \l_tmpa_seq
  \seq_sort:Nn \l_tmpa_seq { \fp_compare:nNn {##2} > {##1} }
  \seq_item:Nn \l_tmpa_seq { #1 }
 }
\ExplSyntaxOff

\begin{document}
\title{RecPFN: Prior-Fitted Networks for In-Context-Based Recommendations}


\author{En Zhi Tan}
\affiliation{%
  \institution{SAP SE}
  \country{Singapore}}
\email{en.zhi.tan@sap.com}

\author{Jia Xiang Lim}
\affiliation{%
  \institution{SAP SE}
  \country{Singapore}}
\email{jia.xiang.lim@sap.com}

\author{Bryan Lijie Chew}
\affiliation{%
  \institution{SAP SE}
  \country{Singapore}}
\email{bryan.chew@sap.com}

\author{Tze Minh Ng}
\affiliation{%
  \institution{SAP SE}
  \country{Singapore}}
\email{tze.minh.ng@sap.com}

\author{Benjamin Yan Han Yap}
\affiliation{%
  \institution{SAP SE}
  \country{Singapore}}
\email{benjamin.yap@sap.com}

\renewcommand{\shortauthors}{En Zhi Tan, Jia Xiang Lim, Bryan Lijie Chew, Tze Minh Ng, and Benjamin
Yan Han Yap}

\begin{abstract} 
We introduce RecPFN, a prior-fitted network that brings in-context learning to sequential recommendation. RecPFN is pretrained entirely on synthetic clickstream environments sampled from a broad structural causal prior, enabling it to amortize Bayesian-style inference from a small support set. At inference, a lightweight decoder-only transformer conditions on a handful of domain sequences and produces next-item predictions for queries in a single forward pass, without any weight updates. Across eight public benchmarks, RecPFN achives state-of-the-art zero-shot performance while remaining strongly competitive with supervised methods in low-compute and low-data regimes. It is deployment-efficient and robust to domain shift, outperforming strong zero-shot baselines that rely on large real-interaction corpora. RecPFN provides a practical path toward generalizable, data-efficient recommenders and opens avenues for richer priors, longer-context ICL, and multimodal extensions. Code for training and evaluation is publicly available at \href{https://github.com/SAP-samples/tabular-ai-recpfn/}{github.com/SAP-samples/tabular-ai-recpfn}.
\end{abstract}



\maketitle

\section{Introduction}

Recommender systems personalize large information spaces to drive engagement and business outcomes \cite{zhu2025ttgl, valencia2024artificial, iana2024survey, sharma2024survey}. The strongest sequential models (e.g., self-attention/transformers) typically require large in-domain datasets, extensive training, and significant online-learning compute \cite{kang2018self, sun2019bert4rec, zhou2020s3, yang2023generic, hebert2025flare, zhai2024actions, yan2025unlocking, chen2024macro}.

Inspired by foundation models in other domains \cite{ramesh2021zero, achiam2023gpt, liu2024sora, liu2024deepseek}, recent work explores generalist recommenders via LLM prompting \cite{lyu2024llm, wang2024whole, liu2025improving} or by pretraining recommendation-specific architectures on diverse user datasets \cite{dingzero, jiang2025recgpt, sheng2025language}. However, these approaches struggle with true zero-shot transfer: preferences and interaction patterns are highly setting-dependent, and mixed-domain pretraining often fails to capture the nuances of a new target environment.

This motivates in-context learning (ICL), where a model adapts at inference by conditioning on a small set of examples without weight updates \cite{dong2024survey}. Beyond LLMs, ICL has proven effective on structured data \cite{ma2024tabdpt, arazitabstar}. Prior-Fitted Networks (PFNs) go further by training on diverse synthetic distributions to approximate Bayesian inference, enabling accurate predictions from few-shot context in a single forward pass \cite{hollmann2022tabpfn, qu2025tabicl, hollmann2025accurate}.

We introduce RecPFN, to our knowledge the first embedding-based PFN-style ICL approach for sequential recommendation. Our contributions are:
\begin{itemize}
\item A synthetic framework for generating clickstream embedding sequences under a broad, expressive prior.
\item A lightweight transformer that consumes example sequences from the target domain and predicts the next item for query sequences in a single pass.
\end{itemize}

Pretraining solely on synthetic data yields strong performance, outperforming zero-shot baselines and rivaling supervised methods with far less compute. RecPFN enables data-efficient, pure in-context adaptation for new domains, offering a practical path toward generalizable recommenders.

\section{Related work}

Sequential recommendation has progressed from RNNs and CNNs to transformers \cite{tan2016improved, zhu2017next, tang2018personalized, kang2018self, sun2019bert4rec, zhou2020s3}. While classical systems are largely ID-based, text and multimodal embeddings improve generalization and cold-start robustness \cite{kanwal2021review, li2023text, sheng2025language, zhou2023mmrec, liu2024multimodal}. Recent approaches integrate pretrained language models, either frozen or fine-tuned for recommendation \cite{li2025llm, kim2024large, liu2025llmemb, bao2025bi}.

The push for foundational recommenders spans two directions. LLM prompting methods format histories and candidates as text, but often face higher latency and difficulty capturing latent collaborative signals \cite{lyu2024llm, wang2024whole, liu2025improving, liang2025taxonomy}. Pretraining recommendation-specific architectures on large, aggregated datasets has also been explored \cite{wang2023pre, wang2024pre, zhang2023hierarchical, li2025llm, dingzero, jiang2025recgpt, sheng2025language}, including domain-invariant learning and cross-domain embedding alignment; yet these domain-agnostic strategies often underperform in new settings due to environment-specific behavioral logic.

ICL offers adaptation without weight updates. Beyond prompting LLMs for ranking/prediction \cite{wang2024whole, yang2025instructional}, PFNs achieve ICL on structured data by training on diverse synthetic tasks and amortizing Bayesian inference \cite{hollmann2022tabpfn}. Our work is the first to extend PFN-style ICL to sequential recommendation.

Synthetic data supports augmentation, privacy, and evaluation in recommenders \cite{antulov2014synthetic, stavinova2022synthetic, yin2024dataset}, via autoencoders, clickstream statistics, or latent preference modeling \cite{shafqat2022hybrid, yan2025synthetic, belletti2019scalable, zhang2021causerec, wang2022causal}. We adopt a causal, latent-factor perspective: a broad prior over a structural causal model defines item properties and transitions, enabling sampling of diverse environments. Training on this distribution teaches RecPFN a general sequential learning algorithm that adapts via in-context examples rather than overfitting to domain-specific idiosyncrasies.

\begin{figure*}
    \centering
    \includegraphics[width=\linewidth]{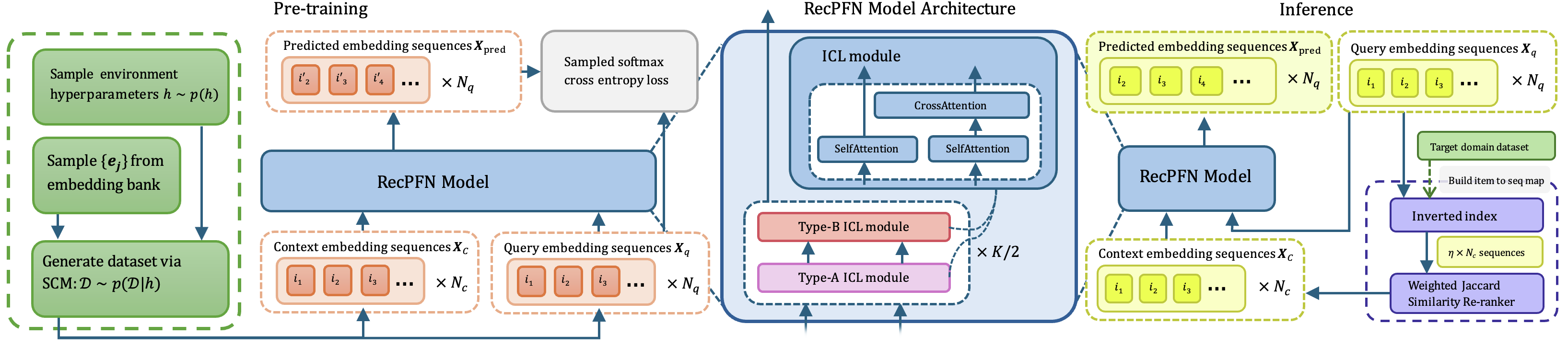}
    \caption{Overall framework of RecPFN: (From left to right) Synthetic data-driven pre-training on next-item prediction task; RecPFN model architecture comprising of alternating Type-A and Type-B ICL modules; Inference based on on-the-fly adaptation to support set retrieved via a recency-weighted overlap}
    \label{fig:overview}
\end{figure*}

\section{RecPFN: In-Context Learning for Sequential Recommendation}
\label{sec:methodology}

In this section, we describe the Bayesian framework underlying prior-fitted networks, the synthetic data generation procedure used for pre-training, and the architecture that enables single-pass next-item prediction conditioned on in-context examples. An overview is shown in Figure \ref{fig:overview}.

\subsection{Bayesian view of prior-fitted networks for sequential recommendation}
\label{ssec:bayes}

Let the item catalog be $\mathcal{I}=\{1,\dots,N\}$ and a user history be $S_{1:t}=(s_1,\dots,s_t)$ with $s_\tau \in \mathcal{I}$. The task is to predict the next item $s_{t+1}\in\mathcal{I}$ given $S_{1:t}$. An environment (hypothesis) $h\in\mathcal{H}$ specifies the sequence-evolution law via $p(j \mid S_{1:t}, h)$ for $j\in\mathcal{I}$. A prior $p(h)$ over $\mathcal{H}$ encodes our beliefs about plausible environments. In RecPFN, $h$ is instantiated by a structural causal model (SCM) and its hyperparameters (Section~\ref{ssec:synth_data}). Training data are sampled by first drawing an environment and then sequences under its dynamics:
\begin{align*}
h &\sim p(h), \;\; \mathcal{D} = \{(S_{1:t}^{(n)}, y^{(n)})\}_{n=1}^{M} \sim p(\mathcal{D}\mid h),
\end{align*}
where $y^{(n)}=s_{t+1}^{(n)}$ is the next item following $S_{1:t}^{(n)}$. At test time in a fixed (but unknown) environment, given an observed dataset $\mathcal{D}$ (e.g., the train split) and a query prefix $S_{1:t}$, the Bayesian posterior predictive distribution over the next item is
\begin{align}
p(j \mid S_{1:t}, \mathcal{D})
&= \int_{\mathcal{H}} p(j \mid S_{1:t}, h)\, p(h \mid \mathcal{D}) \, dh, \quad j\in\mathcal{I}, \label{eq:ppd}\\
p(h \mid \mathcal{D})
&= \frac{p(\mathcal{D}\mid h)\, p(h)}{\int_{\mathcal{H}} p(\mathcal{D}\mid h')\, p(h') \, dh'}. \label{eq:posterior}
\end{align}
Equations \eqref{eq:ppd}–\eqref{eq:posterior} yield the gold-standard target distribution for the next item. We approximate the PPD in \eqref{eq:ppd} with a prior-fitted network $f_{\theta}$ that takes $(S_{1:t}, \mathcal{D})$ and outputs a predictive distribution $\pi_{\theta}(\cdot \mid S_{1:t}, \mathcal{D})$ over $\mathcal{I}$. We optimize $\theta$ by minimizing the expected cross-entropy under the joint prior over environments and data:
\begin{align*}
\mathcal{L}(\theta)
&= \mathbb{E}_{(h,\mathcal{D}_c) \sim p(h)\,p(\mathcal{D}\mid h)}
   \mathbb{E}_{(S_{1:t},y) \sim p(S_{1:t},y \mid h)}  \\
&\quad \big[-\log \pi_{\theta}(y \mid S_{1:t}, \mathcal{D}_c)\big].
\end{align*}
where $\mathcal{D}_c$ is the context (support) set. In practice, the expectations are approximated via Monte Carlo using our synthetic generator by sampling $h \sim p(h)$, then $\mathcal{D} \sim p(\mathcal{D}\mid h)$.

\subsection{Synthetic Sequential Data Generation via a Causal Prior}
\label{ssec:synth_data}

We pre-train RecPFN on synthetic environments sampled from a broad prior over a structural causal model (SCM) of user behavior. At a high level, each environment defines a catalog of $N$ items with embeddings and a transition matrix $\mathbf{T}$ encoding item-to-item affinities. A user sequence is generated by starting from a random item and, at each step, sampling the next item from a mixture of recency-weighted transition scores (capturing short-term sequential patterns) and a global popularity distribution (capturing long-term static appeal). By sampling environments from a broad prior over the SCM hyperparameters, we expose RecPFN to a wide variety of sequence dynamics during pre-training. Algorithm~\ref{alg:sdg} summarizes the full procedure.

\begin{algorithm}[t]
\caption{Synthetic Environment and Sequence Generation}
\label{alg:sdg}
\begin{algorithmic}[1]
\Require Prior ranges for hyperparameters $N, d, \omega, \alpha_{\text{pop}}, \ldots$; embedding bank $\mathcal{E}$
\State Sample hyperparameters: $N, d, \omega, \alpha_{\text{pop}}, \ldots \sim p(h)$
\State Sample $N$ item embeddings $\{\mathbf{e}_j\}_{j=1}^N$ from $\mathcal{E}$
\State Construct transition matrix $\mathbf{T} \in \mathbb{R}^{N \times N}$ via random-graph or latent-factor prior (Section~\ref{ssec:transition_priors})
\State Sample popularity $P_{\text{pop}} \sim \mathrm{Softmax}(\mathcal{N}(0, \sigma^2))$
\For{each sequence $s$ in the batch}
    \State $s_1 \sim \mathrm{Uniform}(\mathcal{I})$
    \For{$t = 2, \ldots, L$}
        \State $\mathbf{p}'_{\text{trans}} \leftarrow \sum_{k=\max(1,t-d)}^{t-1} \omega^{t-1-k}\,\mathbf{T}_{s_k,:}$
        \State $\mathbf{p}_{\text{next}} \propto (1-\alpha_{\text{pop}})\,\mathbf{p}'_{\text{trans}} + \alpha_{\text{pop}}\,P_{\text{pop}}$
        \State $s_t \sim \mathbf{p}_{\text{next}}$ \quad (excluding previously seen items)
    \EndFor
\EndFor
\State \Return sequences, item embeddings $\{\mathbf{e}_j\}$
\end{algorithmic}
\end{algorithm}

\subsubsection{Structural Causal Model for User Behavior}
\label{sss:scm}

Let the item catalog of $N$ items be $\mathcal{I}=\{1,\dots,N\}$. Each item $j$ has an embedding $\mathbf{e}_j \in \mathbb{R}^{D_{\text{item}}}$ and a global popularity $P_{\text{pop},j}$ with $\sum_j P_{\text{pop},j}=1$. Let $\mathbf{T}\in\mathbb{R}^{N\times N}$ be an (unnormalized) transition score matrix; $T_{jk}$ scores the tendency to visit $k$ after $j$.

Given a sequence $S=(s_1,\dots,s_L)$, we generate:
\begin{itemize}
\item Initialization: $s_1 \sim \mathrm{Uniform}(\mathcal{I})$.
\item For $t>1$, form a recency-weighted transition vector from the last $d$ items:
\begin{align}
\mathbf{p}'_{\text{trans}} &= \sum_{k=\max(1,t-d)}^{t-1} w_{t-1-k}\,\mathbf{T}_{s_k,:}, \\
\mathbf{p}_{\text{trans}} &= \frac{\mathbf{p}'_{\text{trans}}}{\sum_j p'_{\text{trans},j}},
\end{align}
with weights $w_\tau \ge 0$. The next item is sampled from
\begin{align}
P(s_t=j \mid s_{<t})
\propto (1-\alpha_{\text{pop}})\,p_{\text{trans},j}+\alpha_{\text{pop}}\,P_{\text{pop},j},
\end{align}
with $\alpha_{\text{pop}} \in [0,1]$ blending short-term sequential context (transition) and long-term static appeal (popularity). The distribution is normalized before sampling.
\end{itemize}

\subsubsection{Transition-Matrix Priors}
\label{ssec:transition_priors}

The sequence dynamics depend critically on the structure of $\mathbf{T}$. We introduce two complementary priors that cover qualitatively different real-world patterns: a \emph{random-graph prior} for sparse, topology-driven transitions (modeling settings where items have a fixed set of natural successors, e.g., series or bundles), and a \emph{latent-factor prior} for factor-structured transitions (modeling settings where co-consumption is driven by shared latent attributes, e.g., genre or category preferences).

\paragraph{Random-Graph Prior.}
This prior models environments where each item has a small set of likely successors defined independently of embedding similarity—capturing niche co-purchase or playlist-style sequences. For each $j\in\mathcal{I}$:
\begin{itemize}
\item Sample a neighbor set $\mathcal{K}_j \subset \mathcal{I}\setminus\{j\}$ of size $k_{\text{rel}}$; set $T_{jk}=1$ for $k\in\mathcal{K}_j$.
\item Set $T_{jj}=1$ with probability $p_{\text{self}}$, else $0$.
\end{itemize}
All other entries are $0$, yielding a sparse, feature-agnostic transition structure.

\paragraph{Latent-Factor Prior.}
This prior models environments where transitions are mediated by shared latent concepts—capturing settings such as genre affinity in media or category co-consumption in retail. Let $D_{\text{latent}}$ be the latent dimension.
\begin{itemize}
\item Sample latent embeddings $\mathbf{L}_{\text{embed}} \in \mathbb{R}^{D_{\text{latent}}\times D_{\text{item}}}$ and a latent transition matrix $\mathbf{T}_{\text{latent}} \in \mathbb{R}^{D_{\text{latent}}\times D_{\text{latent}}}$ (via the random-graph construction).
\item Sample a sparse item–factor map $\mathbf{F}\in\mathbb{R}^{N\times D_{\text{latent}}}$ with entries in $\{-1,1\}$ and about $k_{\text{fac}}$ nonzeros per row.
\item Construct item embeddings and off-diagonal transitions:
\begin{align}
\mathbf{E} &= \mathbf{F}\,\mathbf{L}_{\text{embed}} \in \mathbb{R}^{N\times D_{\text{item}}}, \\
\mathbf{T}' &= \mathbf{F}\,\mathbf{T}_{\text{latent}}\,\mathbf{F}^\top \in \mathbb{R}^{N\times N}.
\end{align}
Set $T_{jk}=T'_{jk}$ for $j\neq k$, and $T_{jj}=1$ w.p.\ $p_{\text{self}}$ (else $0$).
\end{itemize}
Since $\mathbf{L}_{\text{embed}}$ and $\mathbf{T}_{\text{latent}}$ are sampled independently, this decouples feature similarity from sequential transitions. Matrix factorization is a special case corresponding to $\mathbf{T}_{\text{latent}}=\mathbf{I}$.

\subsubsection{Prior over Environments}

We place broad priors over SCM hyperparameters (e.g., $N$, $d$, $w$, $\alpha_{\text{pop}}$, $k_{\text{rel}}$, $p_{\text{self}}$, $D_{\text{latent}}$, $k_{\text{fac}}$). Sampling $h\sim p(h)$ yields diverse environments with varying popularity strength, locality, and factor structure.

\subsubsection{Sampling Embedding Vectors}
\label{ssec:sampling_embeddings}

We build a fixed corpus of 800{,}000 sentence embeddings from BEIR datasets (Quora, FIQA, FEVER, MS~MARCO), encoded with the target LLM. For the random-graph prior, we sample $N$ item embeddings $\{\mathbf{e}_j\}$ from this bank. For the latent-factor prior, we sample $D_{\text{latent}}$ latent vectors for $\mathbf{L}_{\text{embed}}$ and derive item embeddings via $\mathbf{E}=\mathbf{F}\mathbf{L}_{\text{embed}}$. This aligns synthetic embeddings with downstream usage.

\subsubsection{Sampling Procedure}

To draw a training batch: (1) sample an environment $h\sim p(h)$; (2) instantiate item embeddings and $\mathbf{T}$; (3) generate support and query sequences via the SCM. This yields an effectively infinite stream of diverse sequential tasks for pre-training.

\subsection{RecPFN Architecture and In-Context Learning}
\label{ssec:architecture}

RecPFN is a lightweight decoder-only transformer that performs in-context learning by conditioning query sequences on a small set of support sequences from the target domain. The model stacks $K$ custom ICL blocks that first build intra-sequence representations (causal self-attention) and then integrate support information into queries (context-to-query cross-attention). This design enables single-pass next-item prediction without weight updates.

\subsubsection{Input Formulation for In-Context Learning}
\label{sssec:input_formulation}
A prompt consists of $N_c$ support sequences and $N_q$ query sequences, each padded to length $L$ with embedding dimension $D$. The model input can thus be expressed as $\mathbf{X} = (\mathbf{X}_c, \mathbf{X}_q)$ with $\mathbf{X}_c \in \mathbb{R}^{N_c \times L \times D}$ and $\mathbf{X}_q \in \mathbb{R}^{N_q \times L \times D}$.

\subsubsection{Model Architecture}
\label{sssec:model_arch}
Each ICL block comprises:
\begin{itemize}
    \item Intra-sequence self-attention on $\mathbf{X}_c$ and $\mathbf{X}_q$ (with causal masking), producing per-sequence representations.
    \item Context-to-query cross-attention where query tokens attend over the flattened support sequences, enabling query-conditioned retrieval of relevant support information.
\end{itemize}

Let $\text{Attn}(\mathbf{Q}, \mathbf{K}, \mathbf{V})$ denote a multi-head attention operator. Self- and cross-attention on $\mathbf{X} = (\mathbf{X}_c , \mathbf{X}_q)$ can then be denoted as:
\begin{equation*}
    \text{SelfAttention}(\mathbf{X}) = \big(\text{Attn}(\mathbf{X}_{c}, \mathbf{X}_{c}, \mathbf{X}_{c}), \;\text{Attn}(\mathbf{X}_{q}, \mathbf{X}_{q}, \mathbf{X}_{q})\big)
\end{equation*}
\begin{equation*}
    \text{CrossAttention}(\mathbf{X}) = (\mathbf{X}_c , \text{Attn}(\mathbf{X}_{q}, \;\text{flatten}(\mathbf{X}_{c}), \;\text{flatten}(\mathbf{X}_{c})))
\end{equation*}
The $k$-th ICL block updates $(\mathbf{X}^{k-1}_{c}, \mathbf{X}^{k-1}_{q})$ to $(\mathbf{X}^{k}_{c}, \mathbf{X}^{k}_{q})$ by:
\begin{align}
    \mathbf{H}^{(k)} = \text{ICL}(\mathbf{H}^{(k-1)}) \nonumber= \text{CrossAttention}\big(\text{SelfAttention}(\mathbf{H}^{(k-1)})\big)
\end{align}

\subsubsection{Alternating Attention Mechanisms}
\label{sss: attn_blk}
To balance copying-style operations and feature extraction, RecPFN alternates two attention types per block.

\paragraph{Type-A (summed-head attention, no FFN)} The value vector for each head is projected to the full embedding dimension $D$. We then specially initialize $\mathbf{W}^V_i$ to be the identity matrix. Here, we take inspiration from \cite{jelassi2024repeat} who demonstrate powerful and efficient sequence copying abilities with such a formulation. We replace the MLP with a summation across the heads since it has been demonstrated that attention-only transformers are capable of information movement along the sequence \cite{elhage2021mathematical} and defer the learning of nonlinear latent features to Type-B attention.
\begin{align}
\mathbf{Q}_i &= \mathbf{Z} \mathbf{W}^Q_i, \quad \mathbf{K}_i = \mathbf{Z} \mathbf{W}^K_i, \quad \mathbf{V}_i = \mathbf{Z} \mathbf{W}^V_i \\
&\text{where } \mathbf{W}^Q_i, \mathbf{W}^K_i \in \mathbb{R}^{D \times d_k}, \text{ but } \mathbf{W}^V_i \in \mathbb{R}^{D \times D} \nonumber \\
\text{head}_i &= \text{softmax}\left(\frac{\mathbf{Q}_i \mathbf{K}_i^\top}{\sqrt{d_k}} + \mathbf{M}\right) \mathbf{V}_i \in \mathbb{R}^{L' \times D} \\
\text{Attn}_{\text{A}}(\mathbf{Z}) &= \text{LayerNorm}\left(\mathbf{Z} + \sum_{i=1}^{N_h} \text{head}_i\right)
\end{align}

\paragraph{Type-B (standard multi-head attention with FFN)} This is the conventional transformer attention mechanism. The value vector is split across heads, and their outputs are concatenated and passed through a final linear layer and a feed-forward network.
\begin{align}
\mathbf{Q}_i &= \mathbf{Z} \mathbf{W}^Q_i, \quad \mathbf{K}_i = \mathbf{Z} \mathbf{W}^K_i, \quad \mathbf{V}_i = \mathbf{Z} \mathbf{W}^V_i \\ 
&\text{ where } \mathbf{W}^Q_i, \mathbf{W}^K_i, \mathbf{W}^V_i \in \mathbb{R}^{D \times d_k} \text{ and } d_k = D/N_h \nonumber \\
\text{head}_i &= \text{softmax}\left(\frac{\mathbf{Q}_i \mathbf{K}_i^\top}{\sqrt{d_k}} + \mathbf{M}\right) \mathbf{V}_i \in \mathbb{R}^{L' \times d_k} \\
\mathbf{H}_{\text{cat}} &= \text{Concat}(\text{head}_1, \dots, \text{head}_{N_h}) \mathbf{W}^O \\
\mathbf{Z}' &= \text{LayerNorm}(\mathbf{Z} + \mathbf{H}_{\text{cat}}) \\
\text{Attn}_{\text{B}}(\mathbf{Z}) &= \text{LayerNorm}(\mathbf{Z}' + \text{FFN}(\mathbf{Z}'))
\end{align}

\paragraph{Hard-Alibi Masking.} The mask term $\mathbf{M}$ in the attention computation implements hard-alibi masking \cite{jelassi2024repeat}. Given $N_h$ attention heads, for the first half of the heads ($i < N_h/2$), the mask $\mathbf{M}$ is constructed to be highly restrictive, allowing each token to attend only to itself and the preceding $i$ tokens. For the remaining heads ($i \ge N_h/2$), $\mathbf{M}$ applies a standard causal mask.

\subsubsection{Training Objective}
During pre-training on the synthetic data, the model is trained autoregressively to predict the next item's embedding for each position in the query sequences using a sampled softmax loss \cite{wu2024effectiveness}. This forces the model to learn a general algorithm for sequential pattern inference from the provided context.

\subsubsection{Inference on a New Domain} \label{sssec:inference} 

RecPFN adapts in-context without weight updates. The context selector is a structural component of this framework: the quality of in-context adaptation depends directly on the relevance of the retrieved support set, since it is through these sequences that the model conditions on the target domain's behavioral dynamics. From a Bayesian perspective, higher-quality retrieval yields a support set that better approximates a draw from the posterior over environments, directly improving inference accuracy. For a query $S_q=(s_{q,1},\dots,s_{q,L_q})$, we select $N_c$ support sequences from the target domain dataset via a recency-weighted overlap (i.e. sequences containing recent items from the query are prioritized): 
\begin{equation} 
\text{Sim}(S_q, S_j) = \sum_{i \in S_q \cap S_j} w(i, S_q), \quad w(i, S_q) = \lambda^{L_q - t} 
\end{equation} 
Candidate sequences are retrieved efficiently via an inverted index (built over item-to-sequence mappings) starting from the most recent items in the query until some predefined size $\eta \times N_c$ is reached (we set $\eta=10$), then reranked by $\text{Sim}(\cdot,\cdot)$. The top $N_c$ sequences are kept to form ${\mathbf{X}c}$. A single forward pass yields a next-item embedding $\hat{\mathbf{e}}_{pred}$ per query, and final items are retrieved by cosine similarity: 
$\text{Top-K} = \underset{j \in \mathcal{I}_{\text{target}}}{\text{argmax}K} \left( \frac{\hat{\mathbf{e}}_{pred} \cdot \mathbf{e}j}{|\hat{\mathbf{e}}_{pred}| |\mathbf{e}_j|} \right)$.


\section{Experiments}
\label{sec:experiments}

We conduct comprehensive experiments to evaluate RecPFN's effectiveness. Given its single-pass inference and pretraining on varied synthetic data, we expect RecPFN to perform well in low-compute and low-data regimes. Consequently, we evaluate RecPFN against:

\begin{enumerate}
    \item State-of-the-art zero-shot recommendation methods.
    \item Supervised methods trained on the target domain under limited compute.
    \item Supervised methods trained on the target domain with limited data.
\end{enumerate}

\begin{table}
\caption{Dataset statistics}
\label{tab:datasets}
\begin{tabular}{lrrrr}
\toprule
Dataset & Users & Items & Interactions & Ave. Length \\
\midrule
\multicolumn{5}{l}{\textbf{Evaluation}} \\
\midrule
Appliances & 703 & 30,252 & 6,875 & 9.78 \\
Arts & 1,579,230 & 302,809 & 2,875,917 & 1.82 \\
Dianping & 208,596 & 243,247 & 4,422,473 & 21.20 \\
Games & 100,955 & 71,982 & 733,447 & 7.27 \\
Movies & 450,757 & 182,032 & 4,169,631 & 9.25 \\
Pantry & 247,659 & 10,814 & 471,614 & 1.90 \\
Scientific & 28,096 & 165,764 & 199,798 & 7.11 \\
Yelp & 371,243 & 150,346 & 4,728,381 & 12.74 \\
\midrule
\multicolumn{5}{l}{\textbf{Pretraining for ablation experiment}} \\
\midrule
Fashion & 8,886 & 186,189 & 44,769 & 5.04 \\
Grocery & 248,194 & 283,507 & 1,841,452 & 7.42 \\
Movies & 450,757 & 182,032 & 4,169,631 & 9.25 \\
Sports & 6,703,391 & 957,764 & 12,980,837 & 1.94 \\
\bottomrule
\end{tabular}
\end{table}

\begin{table*}
    \centering
    \caption{RecPFN performance against SOTA zero-shot baselines}
    \label{tab:zero_shot}
    \begin{tabular}{lllllllllllllrr}
\toprule
Dataset & \multicolumn{2}{c}{RecPFN} & \multicolumn{2}{c}{RecFormer} & \multicolumn{2}{c}{RecGPT} & \multicolumn{2}{c}{UniSRec} & \multicolumn{2}{c}{VQ-Rec} & \multicolumn{2}{c}{EmbKNN} & \multicolumn{2}{c}{Improv.} \\
\cmidrule(lr){2-3} \cmidrule(lr){4-5} \cmidrule(lr){6-7} \cmidrule(lr){8-9} \cmidrule(lr){10-11} \cmidrule(lr){12-13} \cmidrule(lr){14-15}
@10 & HR & MRR & HR & MRR & HR & MRR & HR & MRR & HR & MRR & HR & MRR & HR & MRR \\
\midrule
Appliances & \textbf{.1915} & \textbf{.1471} & \underline{.1250} & .0808 & .1064 & \underline{.0993} & .0709 & .0340 & \dotuline{.1206} & \dotuline{.0844} & .1064 & .0483 & +53.2\% & +48.1\% \\
Arts & \textbf{.2077} & \textbf{.1485} & \dotuline{.1851} & \underline{.1393} & .0390 & .0372 & .1469 & .1124 & .1293 & .1199 & \underline{.1918} & \dotuline{.1301} & +8.3\% & +6.6\% \\
Games & \textbf{.0673} & \textbf{.0417} & \underline{.0585} & .0283 & .0355 & \underline{.0330} & .0239 & .0116 & .0346 & \dotuline{.0302} & \dotuline{.0471} & .0161 & +15.0\% & +26.6\% \\
Movies & \textbf{.1253} & \textbf{.0708} & \underline{.0875} & \underline{.0431} & .0416 & \dotuline{.0367} & .0354 & .0212 & .0428 & .0366 & \dotuline{.0774} & .0268 & +43.2\% & +64.2\% \\
Pantry & \textbf{.2491} & \textbf{.2031} & .1534 & .1004 & .0599 & .0579 & \dotuline{.2069} & .1780 & .1975 & \dotuline{.1853} & \underline{.2377} & \underline{.1913} & +4.8\% & +6.2\% \\
Scientific & \textbf{.1007} & \textbf{.0730} & \underline{.0827} & \dotuline{.0542} & .0655 & \underline{.0595} & .0431 & .0226 & .0735 & .0482 & \dotuline{.0826} & .0480 & +21.7\% & +22.6\% \\
Software & \textbf{.1882} & \textbf{.1038} & \dotuline{.1418} & \dotuline{.0923} & .1096 & .0912 & .1313 & .0906 & .1229 & \underline{.0930} & \underline{.1448} & .0419 & +30.0\% & +11.6\% \\
Dianping & \textbf{.0574} & \textbf{.0269} & \dotuline{.0103} & \underline{.0088} & .0007 & .0002 & .0016 & .0005 & .0003 & .0002 & \underline{.0220} & \dotuline{.0040} & +160.9\% & +204.7\% \\
Yelp & \underline{.0245} & \dotuline{.0165} & .0163 & .0092 & .0179 & \underline{.0171} & .0073 & .0044 & \dotuline{.0186} & \textbf{.0172} & \textbf{.0293} & .0152 & -16.4\% & -4.2\% \\
\bottomrule
\end{tabular}
\par\smallskip\footnotesize Notes: \textbf{Bold} = best, \underline{Underlined} = second-best, \dotuline{Dotted underline} denotes third-best.
\end{table*}

\subsection{Experimental Setup}

\subsubsection{Datasets}
We evaluate on eight public benchmarks for sequential recommendation (Table~\ref{tab:datasets}).

The evaluation suite spans: (i) Amazon Review categories in e-commerce (Appliances; Arts, Crafts and Sewing; Movies; Video Games; Industrial and Scientific; Software), (ii) Yelp for local businesses and services (reviews and check-ins), and (iii) Dianping for O2O services in the Chinese market, which differs linguistically and culturally from Yelp and Amazon. Together these datasets cover distinct platforms (marketplaces vs.\ local services), heterogeneous item semantics, and a wide range of scales and sparsity levels.

For the pre-training-on-real-data ablation (Section~\ref{sss: syn data ablation}), we replace the synthetic prior with four Amazon categories—Fashion, Grocery and Gourmet Food, Sports and Outdoors, and Movies—used only for pre-training. Movies does not appear among the evaluation datasets, avoiding leakage.

\subsubsection{Baselines}
We compare RecPFN with strong supervised and zero-shot baselines. The supervised set includes GRU4Rec \cite{hidasi2015session}, SASRec \cite{kang2018self}, and BERT4Rec \cite{sun2019bert4rec} (ID-based), FDSA \cite{zhang2019feature} (embedding-based), semantic variants of the three ID-based models, and pre-trained recommenders UniSRec \cite{hou2022towards} and VQ-Rec \cite{hou2023learning} fine-tuned on the target domain. Semantic variants replace the discrete item-ID inputs of ID-based models with the same sentence embeddings used by RecPFN, enabling a direct comparison of the ID-based and embedding-based paradigms under identical training conditions. The foundation/zero-shot set includes RecGPT \cite{jiang2025recgpt} as well as zero-shot variants of UniSRec, VQ-Rec, and RecFormer \cite{li2023text}. We additionally include EmbKNN as a zero-shot baseline: given a query sequence, EmbKNN retrieves the top-$K$ items by cosine similarity between the query's most recent item embedding and all catalog item embeddings, without any learned model. This suite spans ID-only, metadata-aware, semantic, LLM-prompted, and pre-trained paradigms.

\begin{table*}
    \centering
    \caption{RecPFN performance against supervised baselines trained for 3 epochs on the full target domain dataset, illustrating deployment efficiency under a per-domain training budget. $^{\wedge}$ denotes the semantic embedding variant of the ID-based models. $^*$ indicates statistically significant improvements (p<0.05).}
    \label{tab:low_compute}
    \begin{tabular}{llllllllllllrr}
\toprule
Dataset & @10 & RecPFN & BERT & BERT$^{\wedge}$ & FDSA & GRU & GRU$^{\wedge}$ & SAS & SAS$^{\wedge}$ & UniSRec & VQ-Rec & Improv. & Rank \\
\midrule
\multirow{2}{*}{Appliances} & HR & \textbf{.1915}$^{*}$ & .0391 & .0938 & .0836 & .0240 & .0000 & .0697 & .1192 & \underline{.1259} & \dotuline{.1242} & +52.1\% & 1 \\
 & MRR & \textbf{.1471}$^{*}$ & .0095 & .0392 & .0235 & .0113 & .0000 & .0655 & .0831 & \dotuline{.0937} & \underline{.0982} & +49.8\% & 1 \\
\midrule
\multirow{2}{*}{Arts} & HR & \textbf{.2077}$^{*}$ & .0396 & .0163 & .0661 & .0264 & .0149 & .0229 & .0484 & \underline{.1936} & \dotuline{.1280} & +7.3\% & 1 \\
 & MRR & \underline{.1485} & .0181 & .0076 & .0458 & .0135 & .0090 & .0094 & .0343 & \textbf{.1531} & \dotuline{.1183} & -3.0\% & 2 \\
\midrule
\multirow{2}{*}{Games} & HR & \textbf{.0673}$^{*}$ & .0372 & .0122 & \dotuline{.0396} & .0244 & .0152 & .0169 & .0352 & \underline{.0636} & .0343 & +5.9\% & 1 \\
 & MRR & \textbf{.0417}$^{*}$ & .0144 & .0053 & .0165 & .0093 & .0099 & .0070 & .0283 & \underline{.0346} & \dotuline{.0300} & +20.5\% & 1 \\
\midrule
\multirow{2}{*}{Movies} & HR & \textbf{.1253}$^{*}$ & .1022 & .0191 & \dotuline{.1078} & .0984 & .0262 & .0227 & .0064 & \underline{.1167} & .0425 & +7.4\% & 1 \\
 & MRR & \dotuline{.0708} & .0674 & .0065 & \textbf{.0814} & .0638 & .0166 & .0083 & .0038 & \underline{.0743} & .0365 & -13.0\% & 3 \\
\midrule
\multirow{2}{*}{Pantry} & HR & \textbf{.2491}$^{*}$ & .0292 & .0340 & .0488 & .0299 & .0577 & .0369 & .0243 & \underline{.2360} & \dotuline{.1994} & +5.5\% & 1 \\
 & MRR & \textbf{.2031}$^{*}$ & .0096 & .0140 & .0197 & .0107 & .0346 & .0149 & .0132 & \underline{.2006} & \dotuline{.1879} & +1.3\% & 1 \\
\midrule
\multirow{2}{*}{Scientific} & HR & \textbf{.1007}$^{*}$ & .0491 & .0425 & .0527 & .0414 & .0352 & .0488 & .0441 & \underline{.0819} & \dotuline{.0719} & +23.0\% & 1 \\
 & MRR & \textbf{.0730}$^{*}$ & .0275 & .0298 & .0298 & .0212 & .0250 & .0265 & .0325 & \underline{.0505} & \dotuline{.0488} & +44.5\% & 1 \\
\midrule
\multirow{2}{*}{Software} & HR & \textbf{.1882} & .0488 & .0596 & .0763 & .0461 & .0509 & .1018 & .1160 & \underline{.1816} & \dotuline{.1211} & +3.6\% & 1 \\
 & MRR & \underline{.1038} & .0146 & .0229 & .0344 & .0144 & .0403 & .0815 & .0740 & \textbf{.1042} & \dotuline{.0920} & -0.4\% & 2 \\
\midrule
\multirow{2}{*}{Dianping} & HR & \textbf{.0574}$^{*}$ & \underline{.0154} & .0004 & .0039 & \dotuline{.0106} & .0001 & .0011 & .0000 & .0068 & .0003 & +273.3\% & 1 \\
 & MRR & \textbf{.0269}$^{*}$ & \underline{.0049} & .0001 & .0012 & \dotuline{.0033} & .0000 & .0003 & .0000 & .0021 & .0002 & +448.0\% & 1 \\
\midrule
\multirow{2}{*}{Yelp} & HR & .0245 & \textbf{.0358} & .0066 & \dotuline{.0293} & \underline{.0356} & .0049 & .0191 & .0019 & .0272 & .0181 & -31.5\% & 5 \\
 & MRR & \underline{.0165} & .0122 & .0023 & .0105 & \dotuline{.0124} & .0019 & .0066 & .0008 & .0124 & \textbf{.0168} & -2.1\% & 2 \\
\bottomrule
\end{tabular}
\par\smallskip\footnotesize Notes: \textbf{Bold} = best, \underline{Underlined} = second-best, \dotuline{Dotted underline} denotes third-best. BERT = BERT4Rec, SAS = SASRec, GRU = GRU4Rec.
\end{table*}

\begin{table*}
    \centering
    \caption{RecPFN performance against supervised baselines trained for 50 epochs on 10\% of the target domain dataset. $^{\wedge}$ denotes the semantic embedding variant of the ID-based models. $^*$ indicates statistically significant improvements (p<0.05)}
    \label{tab:low_data}
    \begin{tabular}{llllllllllllr}
\toprule
Dataset & @10 & RecPFN & BERT & BERT$^{\wedge}$ & FDSA & GRU & GRU$^{\wedge}$ & SAS & SAS$^{\wedge}$ & UniSRec & VQ-Rec & Rank \\
\midrule
\multirow{2}{*}{Appliances} & HR & \textbf{.1348} & .0868 & .0834 & .0960 & .0295 & .0428 & .1221 & \underline{.1342} & \dotuline{.1277} & .1277 & 1 \\
 & MRR & \textbf{.1016}$^{*}$ & .0225 & .0513 & .0314 & .0100 & .0214 & .0632 & .0829 & \underline{.0931} & \dotuline{.0887} & 1 \\
\midrule
\multirow{2}{*}{Arts} & HR & \underline{.1947} & .0531 & .0173 & .1102 & .0720 & .0157 & .0973 & .0658 & \textbf{.1990} & \dotuline{.1834} & 2 \\
 & MRR & \dotuline{.1424} & .0360 & .0060 & .0920 & .0544 & .0096 & .0716 & .0534 & \underline{.1610} & \textbf{.1656} & 3 \\
\midrule
\multirow{2}{*}{Pantry} & HR & \textbf{.2364} & .0258 & .0296 & .0755 & .0398 & .0405 & .0688 & .0594 & \underline{.2357} & \dotuline{.2187} & 1 \\
 & MRR & \underline{.1990} & .0170 & .0169 & .0583 & .0145 & .0254 & .0506 & .0503 & \textbf{.1993} & \dotuline{.1944} & 2 \\
\midrule
\multirow{2}{*}{Scientific} & HR & \underline{.0865} & .0408 & .0606 & .0785 & .0309 & .0322 & .0626 & .0748 & \textbf{.0936} & \dotuline{.0832} & 2 \\
 & MRR & \underline{.0650} & .0279 & .0488 & \textbf{.0653} & .0226 & .0237 & .0368 & \dotuline{.0646} & .0630 & .0551 & 2 \\
\midrule
\multirow{2}{*}{Dianping} & HR & \textbf{.0275}$^{*}$ & .0054 & .0003 & \underline{.0136} & .0058 & .0001 & \dotuline{.0086} & .0001 & .0069 & .0001 & 1 \\
 & MRR & \textbf{.0074}$^{*}$ & .0018 & .0001 & \underline{.0055} & .0021 & .0000 & \dotuline{.0022} & .0000 & .0019 & .0000 & 1 \\
\midrule
\multirow{2}{*}{Yelp} & HR & .0191 & .0137 & .0068 & \underline{.0324} & .0124 & .0041 & \dotuline{.0258} & .0025 & .0224 & \textbf{.0363} & 5 \\
 & MRR & .0107 & .0044 & .0023 & .0114 & .0041 & .0016 & \dotuline{.0122} & .0006 & \underline{.0148} & \textbf{.0199} & 5 \\
\bottomrule
\end{tabular}
\par\smallskip\footnotesize Notes: \textbf{Bold} = best, \underline{Underlined} = second-best, \dotuline{Dotted underline} denotes third-best. BERT = BERT4Rec, SAS = SASRec, GRU = GRU4Rec.
\end{table*}

\subsubsection{Evaluation Settings}
We report HR@10 and MRR@10 using the full item catalog as the candidate set. Data are split 70/10/20 into train/validation/test; validation and test are each capped at 10{,}000 samples, with the remainder assigned to training. All methods use identical splits. For fine-tuning of supervised baselines, we repeat each experiment across 4 random seeds and average the results.

\subsubsection{Implementation Details}

\paragraph{Architecture.}
RecPFN uses $K{=}6$ stacked ICL modules (Section~\ref{sssec:model_arch}) with alternating Type-A and Type-B attention blocks (Section~\ref{sss: attn_blk}), $N_h{=}8$ attention heads, and dropout of 0.2 on attention and residual connections.

\paragraph{Training.}
Pre-training is performed on a single A100 GPU for 36 hours in two stages: first on the random-graph prior, then on an equal mix of random-graph and latent-factor priors. Each stage runs up to 120 epochs; per epoch we execute 500 training and 200 validation steps. Each step samples one environment and constructs a batch with $N_c{=}128$ support sequences and $N_q{=}16$ query sequences. We use AdamW \cite{loshchilov2017fixing} with a learning rate of $1{\times}10^{-4}$, gradient accumulation of 2, and 6 warm-up epochs. Early stopping (patience 20) monitors the ratio $\text{MRR@10}_{\text{RecPFN}} / \text{MRR@10}_{\text{EmbKNN}}$ on the synthetic validation stream to normalize for environment difficulty. Item embeddings are produced by \texttt{gte-Qwen2-1.5B-instruct} \cite{li2023towards}. Additional details on synthetic data generation appear in Appendix~\ref{apdx: sdg}.

\paragraph{Evaluation.} For evaluation, we use a batch size of 1 with a context size of 8. We set context selector hyperparameters $\lambda=e$ and $\eta=10$.

\paragraph{Baselines.}
SASRec, GRU4Rec, BERT4Rec, and FDSA are implemented via RecBole \cite{xu2023towards} with its optimal hyperparameters. All other baselines use their official repositories and default hyperparameters for fine-tuning (where applicable) and inference.

\subsection{Main Results} \label{ssec:main_results}

We evaluate in three settings:

\subsubsection{Zero-shot (Table~\ref{tab:zero_shot}).} RecPFN exceeds all zero-shot baselines on all but one dataset, despite being pre-trained solely on synthetic data, whereas competing methods rely on large real interaction corpora. Notably, EmbKNN is often competitive—sometimes matching or beating pre-trained baselines—which underscores the strong context dependence of recommendation and calls into question the effectiveness of domain-agnostic pretraining. RecPFN addresses this by performing single-pass inference while adapting on-the-fly via a small, relevant support set, without weight updates.

\subsubsection{Deployment Efficiency (Table~\ref{tab:low_compute}).} We evaluate under a budget of 3 training epochs, where supervised baselines receive multiple full passes over the target-domain data. RecPFN, which never updates weights and performs a single forward pass conditioned on a small support set drawn from the train split, still outperforms most baselines across datasets, including pre-trained UniSRec and VQ-Rec after brief fine-tuning. This comparison reflects RecPFN’s core deployment advantage: it trades a one-time pretraining cost (36 hours on a single A100 GPU) for zero per-domain retraining cost across all future target domains. This amortization becomes increasingly favorable as the number of deployment domains grows.

\subsubsection{Low-data (Table~\ref{tab:low_data}). \label{sss: low_data}} When the train split is truncated to 10\%, both the supervised training signal and RecPFN’s context pool shrink. Pre-trained methods become strongest, while ID-based approaches degrade most. Even after 50 epochs of supervised fine-tuning, RecPFN remains competitive, indicating that its in-context learning more effectively exploits severely limited target data for adaptation than conventional fine-tuning.

\subsection{Performance versus Training Compute}
\label{ssec:epoch_scaling}

We measure MRR@10 versus epochs for SASRec, FDSA, and VQ-Rec on four representative benchmarks, training each supervised baseline on the full split for up to 100 epochs under identical settings, and evaluating after every epoch. RecPFN needs no target-domain training and thus appears as a horizontal line.

\begin{figure}[t]
    \centering
    \includegraphics[width=\linewidth]{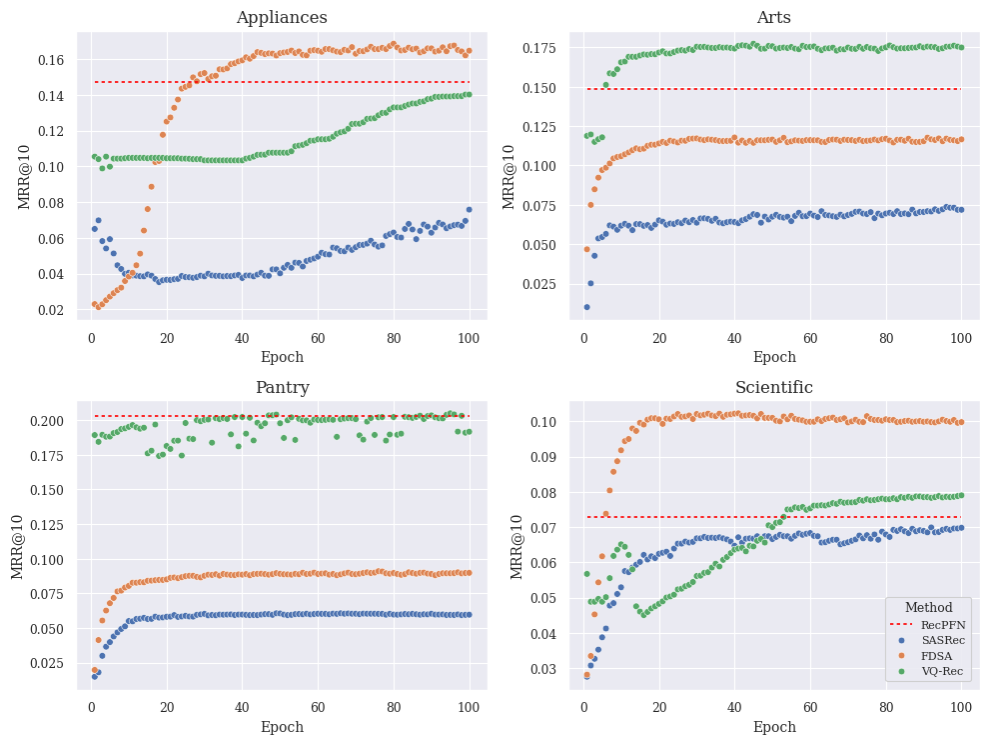}
    \caption{MRR@10 versus training epochs (up to 100) on four datasets for SASRec, FDSA, VQ-Rec, and RecPFN. RecPFN requires no target-domain training and is shown as a horizontal line.}
    \label{fig:epoch_mrr}
\end{figure}

Figure~\ref{fig:epoch_mrr} shows RecPFN remains competitive across the entire compute range. Even after 100 epochs, no supervised method outperforms RecPFN on more than two of the four datasets, and matching RecPFN typically requires substantial per-domain compute. In the early training regime (first 10–20 epochs), RecPFN often leads, making it attractive for rapid deployment and settings with frequent domain shifts where per-domain retraining is costly. These observations confirm that RecPFN achieves strong performance without domain-specific training while supervised methods require significant per-domain compute to match it.

\section{Analysis and Discussion}
\label{sec:analysis}

\begin{table*}
    \centering
    \caption{Relative performance of various ablation experiments for RecPFN (expressed as percentage change)}
    \label{tab:ablation}
    \begin{tabular}{lllllllllllll}
\toprule
\multicolumn{1}{l}{Experiment} & \multicolumn{2}{c}{Appliances} & \multicolumn{2}{c}{Arts} & \multicolumn{2}{c}{Pantry} & \multicolumn{2}{c}{Scientific} & \multicolumn{2}{c}{Dianping} & \multicolumn{2}{c}{Yelp} \\
\cmidrule(lr){2-3} \cmidrule(lr){4-5} \cmidrule(lr){6-7} \cmidrule(lr){8-9} \cmidrule(lr){10-11} \cmidrule(lr){12-13}
@10 & HR & MRR & HR & MRR & HR & MRR & HR & MRR & HR & MRR & HR & MRR \\
\midrule
RecPFN-base & .1915 & .1471 & .2077 & .1485 & .2491 & .2031 & .1007 & \underline{.0730} & .0574 & .0269 & .0245 & .0165 \\
\arrayrulecolor{black}\cmidrule[0.2pt](lr){1-13}
\multicolumn{13}{l}{\textbf{Model}} \\
\arrayrulecolor{black}\cmidrule[0.2pt](lr){1-13}
RecPFN-small & \dotuline{+3.7} & -26.6 & +0.5 & +0.7 & -1.0 & +0.3 & +1.9 & -1.0 & -17.8 & -29.2 & -9.0 & -10.0 \\
RecPFN-mini & +3.7 & -12.8 & +0.5 & -0.8 & -1.3 & +0.2 & \underline{+4.8} & -14.7 & -25.1 & -40.7 & +2.0 & \dotuline{+4.5} \\
\arrayrulecolor{Grey0!60!RoyalBlue3}\cmidrule[0.1pt](lr){1-13}
baai/bge-m3 & -14.8 & -20.9 & \textbf{+2.5} & +0.4 & \textbf{+1.2} & \dotuline{+1.1} & \dotuline{+4.4} & -4.6 & \underline{+17.6} & \dotuline{+16.2} & +2.0 & +0.1 \\
embeddinggemma-300m & +3.7 & -18.0 & -2.0 & -0.9 & -0.7 & +0.4 & +4.2 & -12.9 & \textbf{+24.0} & \textbf{+31.0} & -64.9 & -72.8 \\
multilingual-mpnet-base-v2 & -11.1 & -32.8 & -6.6 & -2.0 & -3.4 & -1.1 & -15.4 & -27.0 & -69.7 & -88.3 & -44.9 & -52.3 \\
\arrayrulecolor{Grey0!60!RoyalBlue3}\cmidrule[0.1pt](lr){1-13}
only type-A attention & +0.0 & -19.7 & +0.5 & +0.8 & -0.3 & +0.5 & +0.0 & -3.0 & -7.0 & -12.3 & -5.3 & -3.7 \\
only type-B attention & -18.5 & -22.4 & -5.8 & -5.7 & -0.4 & -1.4 & \textbf{+6.5} & -12.6 & -61.0 & -83.0 & \textbf{+8.6} & \textbf{+18.4} \\
\arrayrulecolor{black}\cmidrule[0.2pt](lr){1-13}
\multicolumn{13}{l}{\textbf{Synthetic Data}} \\
\arrayrulecolor{black}\cmidrule[0.2pt](lr){1-13}
amazon dataset & -14.8 & -37.5 & -2.5 & -3.0 & -0.3 & -0.1 & -7.1 & -26.0 & -42.9 & -59.3 & -13.9 & -25.7 \\
\arrayrulecolor{Grey0!60!RoyalBlue3}\cmidrule[0.1pt](lr){1-13}
only random graph prior & +0.0 & -11.3 & \underline{+0.8} & \dotuline{+1.5} & -0.4 & +0.1 & -0.2 & -1.5 & -19.5 & -32.2 & -12.2 & -10.7 \\
\arrayrulecolor{black}\cmidrule[0.2pt](lr){1-13}
\multicolumn{13}{l}{\textbf{Inference}} \\
\arrayrulecolor{black}\cmidrule[0.2pt](lr){1-13}
batch size = 2 & -4.4 & -15.0 & -14.1 & -11.6 & -11.2 & -9.4 & -8.1 & -6.5 & -43.6 & -64.0 & -18.8 & -19.8 \\
batch size = 4 & -20.2 & -21.3 & -28.4 & -24.6 & -17.9 & -14.4 & -14.1 & -11.8 & -59.8 & -78.2 & -29.4 & -33.8 \\
batch size = 8 & -16.6 & -18.0 & -30.9 & -27.4 & -16.3 & -12.0 & -20.4 & -17.1 & -65.5 & -84.0 & -39.6 & -43.4 \\
\arrayrulecolor{black}\cmidrule[0.2pt](lr){1-13}
\multicolumn{13}{l}{\textbf{Context}} \\
\arrayrulecolor{black}\cmidrule[0.2pt](lr){1-13}
no context & -74.1 & -86.3 & -69.3 & -71.5 & -38.2 & -44.4 & -82.0 & -85.2 & -69.3 & -87.9 & -86.9 & -89.8 \\
random context & -59.3 & -70.8 & -64.9 & -72.7 & -39.4 & -47.1 & -67.5 & -66.9 & -71.3 & -89.8 & -88.2 & -94.5 \\
\arrayrulecolor{Grey0!60!RoyalBlue3}\cmidrule[0.1pt](lr){1-13}
context size = 2 & +0.0 & \dotuline{+4.2} & +0.5 & +0.2 & -0.7 & +0.2 & +3.4 & -4.2 & +1.7 & +5.2 & \underline{+5.7} & +2.7 \\
context size = 4 & +3.7 & +0.5 & \dotuline{+0.8} & +0.1 & -0.1 & +0.1 & +2.8 & -0.6 & +5.2 & +10.8 & +2.0 & \underline{+5.2} \\
context size = 16 & +3.7 & -4.4 & +0.0 & -0.5 & -0.4 & -0.1 & -1.6 & -2.3 & -15.5 & -25.0 & -2.4 & -6.8 \\
context size = 32 & -3.7 & -9.1 & -0.9 & -1.0 & -0.7 & -0.2 & -1.6 & -2.5 & -39.2 & -57.7 & -7.8 & -17.1 \\
\arrayrulecolor{Grey0!60!RoyalBlue3}\cmidrule[0.1pt](lr){1-13}
10\% context selection dataset & -29.6 & -30.9 & -6.3 & -4.1 & -5.1 & -2.0 & -14.1 & -10.9 & -52.1 & -72.3 & -22.0 & -35.1 \\
\arrayrulecolor{Grey0!60!RoyalBlue3}\cmidrule[0.1pt](lr){1-13}
$\lambda=1$ (unweighted similarity score) & +0.0 & -7.8 & -0.5 & -0.3 & -0.8 & -0.3 & -1.1 & -0.4 & -36.8 & -52.1 & -5.7 & -11.4 \\
$\lambda=e^2$ & -3.7 & +0.2 & -0.5 & -0.2 & \underline{+0.3} & +0.0 & +0.9 & \dotuline{-0.2} & +1.2 & -0.8 & +0.0 & +1.2 \\
\arrayrulecolor{Grey0!60!RoyalBlue3}\cmidrule[0.1pt](lr){1-13}
$\eta=3$ & -3.7 & -0.6 & -0.6 & -0.3 & -0.3 & +0.1 & -0.4 & -1.5 & -36.9 & -52.7 & -9.4 & -11.9 \\
$\eta=50$ & +0.0 & +1.0 & -0.2 & -0.1 & \dotuline{+0.1} & +0.0 & +0.0 & \textbf{+0.2} & \dotuline{+12.4} & \underline{+23.5} & \dotuline{+2.4} & +2.3 \\
\bottomrule
\end{tabular}
\par\smallskip\footnotesize Notes: \textbf{Bold} = best, \underline{Underlined} = second-best, \dotuline{Dotted underline} denotes third-best.

\end{table*}

\subsection{Ablation Studies \label{ss: ablation}}

We analyze RecPFN across four axes: (1) model design, (2) synthetic data priors, (3) inference-time batching, and (4) context construction. Overall, the ablations support our design choices and highlight where further gains are possible.

\subsubsection{Model Ablations}

\paragraph{Model size.}
We evaluate two reduced-capacity variants—RecPFN-small (4 ICL blocks) and RecPFN-mini (2 ICL blocks). Both remain competitive overall but show regressions on some datasets, suggesting that depth helps RecPFN learn a stronger in-context algorithm. Increasing the difficulty and diversity of the synthetic prior (e.g., longer-range dependencies, multi-step motifs, hierarchical patterns) should better saturate capacity and further separate the larger model from smaller variants.

\paragraph{Language model.}
We train RecPFN with \texttt{BGE M3} \cite{chen2024bge}, \texttt{Embedding Gemma-300M} \cite{vera2025embeddinggemma}, and \texttt{paraphrase-multilingual-mpnet-base-v2} \cite{reimers-2019-sentence-bert}. The multilingual model is included to probe cross-lingual robustness, given that our evaluation suite includes Dianping, a Chinese-market dataset. Smaller encoders cause noticeable regressions on some datasets, but overall performance remains solid, indicating robustness to moderate embedding shifts. Stronger embedding geometry directly benefits recommendation quality, since inference relies on cosine retrieval over the item catalog (Section~\ref{sssec:inference}).

\paragraph{Attention mechanisms.}
Variants using only Type-A or only Type-B attention (Section~\ref{sss: attn_blk}) underperform the alternating design. Type-B-only is markedly worse, supporting that Type-A is crucial for efficient copying and sequence-conditioned retrieval over flattened context. Alternation balances feature extraction (Type-B) and copying-style operations (Type-A), stabilizing training and yielding the best generalization.

\subsubsection{Synthetic Data \label{sss: syn data ablation}}

\paragraph{Pre-training on real-world data.}
Replacing the synthetic SCM prior with large-scale real datasets (Amazon Fashion, Grocery, Movies, Sports) significantly lowers performance across evaluation domains, consistent with our claim that recommendation is highly context-dependent: domain-agnostic pre-training on real interactions fails to generalize zero-shot, whereas the synthetic prior trains amortized inference rather than memorizing domain idiosyncrasies.

\paragraph{Prior composition.}
Training exclusively on the random-graph prior is generally poorer than the mixed-prior setting, indicating that the latent-factor prior is essential for capturing factor-driven co-consumption and non-trivial item–item dependencies. Conversely, training solely on the latent-factor prior did not converge reliably due to its higher difficulty, motivating our two-stage curriculum: learn robust sequence-processing primitives on the random-graph prior, then introduce latent-factor dynamics without destabilizing training.

\subsubsection{Inference: Batch size} Scaling batch size and context proportionally degrades performance, suggesting that cross-attention over broad contexts dilutes attention budgets and increases retrieval noise. Improving robustness at larger contexts is a promising direction—e.g., hierarchical context indexing, locality-aware routing, gating, or continued pre-training on longer-context regimes.

\subsubsection{Context} \paragraph{Context size.}
Removing the support set causes drastic drops across all datasets, confirming that RecPFN’s gains stem from genuine in-context adaptation. Random context also causes drastic regression, showing context does not merely have a regularization effect. Small context sizes remain close to peak, while larger contexts introduce slight degradations, likely from reduced signal-to-noise.

\paragraph{Limited context-selection pool.}
Constraining the candidate pool for context selection causes significant degradation, consistent with the need for a sufficiently large pool to find high-similarity exemplars. Nevertheless, as in our low-data experiments (Section \ref{sss: low_data}), RecPFN remains competitive against fully trained supervised baselines when both training data and context pool are limited.

\paragraph{Context selector.}
Replacing the temporally weighted similarity with an unweighted Jaccard-style overlap degrades performance across all datasets, underscoring the importance of recency and order - consistent with our sequence-centric synthetic generation (Section \ref{ssec:synth_data}) and hard-alibi masking (Section~\ref{sss: attn_blk}). Increasing the decay rate does not significantly impact performance. However, reducing the candidate pool degrades performance, particularly for the largest datasets Dianping and Yelp while increasing it yields significant gains for these datasets.

\paragraph{Remarks on context ablations.}
Together, these ablations establish the context selector as a structural bottleneck: removing context, degrading retrieval quality, or shrinking the candidate pool all cause substantial drops, while the model is insensitive to moderate retrieval noise. This asymmetry reflects the Bayesian framing: the model exploits whatever signal is in the support set, but cannot compensate for its absence.

\subsection{Efficiency Analysis}
\label{ssec:efficiency}

Table~\ref{tab:runtimes} shows the single epoch training runtimes as well as the inference time per user for the best performing baselines from the different categories. For inference, the runtimes are obtained by averaging across the full test split with batch size of 96 for the baselines and 8 for RecPFN. We additionally record preprocessing runtimes for RecPFN (building context selector) and VQ-Rec (building item codes). We omit embedding generation for all algorithms.

RecPFN requires a single pre-training phase and no domain-specific fine-tuning, so deployment compute is dominated by inference. While its inference time is noticeably higher than that of compact supervised baselines, it remains far more efficient than competitive zero-shot methods. This makes RecPFN particularly attractive in settings with frequent distribution shifts where retraining would otherwise be required.

\begin{table}
    \centering
    \caption{RecPFN runtimes in seconds against baselines for Pantry and Yelp. Times reported for single epoch training, inference per user (averaged across full test split)}
    \label{tab:runtimes}
    \begin{tabular}{lrrr}
\toprule
Method & Preprocessing (s) & Training (s) & Inference (ms) \\
\midrule
\multicolumn{4}{l}{\textbf{Yelp}} \\
\midrule
RecPFN & 46.56 & - & 6.58 \\
RecGPT & - & - & 40.81 \\
VQ-Rec & 691.20 & 380.69 & 0.44 \\
FDSA & - & 1016.30 & 2.49 \\
\midrule
\multicolumn{4}{l}{\textbf{Scientific}} \\
\midrule
RecPFN & 10.05 & - & 5.59 \\
RecGPT & - & - & 37.35 \\
VQ-Rec & 401.94 & 199.32 & 0.46 \\
FDSA & - & 44.17 & 1.61 \\
\bottomrule
\end{tabular}
\end{table}

\subsection{Qualitative Analysis: In-Context Adaptation in Action}
\label{ssec:qualitative}

\subsubsection{Case Study on Amazon Movies example}

We illustrate how RecPFN leverages in-context examples to make semantically aligned and sequence-consistent predictions. Table~\ref{tab:qual_case} presents a representative case from Amazon Movies where the ground-truth next item is a faith/family film. RecPFN, conditioning on a small set of retrieved support sequences (Table~\ref{tab:qual_support}), correctly predicts the next item; ablating the context or comparing against EmbKNN demonstrates the importance of in-context adaptation.\footnote{For clarity, predictions shown in Table~\ref{tab:qual_case} exclude items already present in the user's history.}

\begin{table}[t]
\centering
\footnotesize
\caption{Qualitative example (Amazon Movies). RecPFN correctly predicts the ground-truth next item when conditioned on the retrieved support sequences; removing the context or using EmbKNN yields plausible but incorrect recommendations.}
\label{tab:qual_case}
\begin{tabular}{p{0.2\linewidth} p{0.72\linewidth}}
\toprule
Query history &

Last Vegas; The Big Wedding (Digital); Life of Pi; Poverty, Inc.;
Intolerable Cruelty; In the Arms of Angels
 \\
\midrule
Ground truth &
\makecell[l]{A Pioneer Miracle} \\
\midrule
RecPFN Top-3 & A Pioneer Miracle; The Teacher; Savior \\
\midrule
RecPFN (no context) Top-3 & Panic in the Streets (VHS); Frightworld; A Tale of Two Cities \\
\midrule
EmbKNN Top-3 & Silver Linings Playbook; Movies He'll Love; City by the Sea \\
\midrule
Rationale & RecPFN observes support sequences containing the transition: 
In the Arms of Angels $\rightarrow$ A Pioneer Miracle (Table~\ref{tab:qual_support}).
This in-context signal steers the model toward the correct next item.
Without context, predictions revert to broadly similar or popular items. \\
\bottomrule
\end{tabular}
\end{table}

\begin{table}[t]
\centering
\footnotesize
\caption{Most influential support sequences retrieved by the context selector (titles shown). These sequences contain the co-occurrence and transition indicative of the ground-truth next item.}
\label{tab:qual_support}
\begin{tabular}{p{0.03\linewidth} p{0.90\linewidth}}
\toprule
1 &
Secrets of Archeology: The Lost Cities of the Maya; Seven Girlfriends; In the Arms of Angels; A Pioneer Miracle \\
\midrule
2 & About Miracles; Doctor Thorne (Season 1); In the Arms of Angels; A Pioneer Miracle \\
\midrule
3 & In the Arms of Angels; A Pioneer Miracle; Touched By Grace; Before All Others \\
\midrule
4 & Dying of the Light; In the Arms of Angels; Finding Neverland; A Pioneer Miracle \\
\bottomrule
\end{tabular}
\end{table}

This case highlights the mechanism by which RecPFN adapts: the query attends over a compact, high-similarity support set that encodes domain-specific sequential logic (e.g., consistent co-consumption of faith/family titles and the observed transition from \emph{In the Arms of Angels} to \emph{A Pioneer Miracle}). Baselines that lack such on-the-fly conditioning either drift toward globally popular titles or produce semantically plausible but context-mismatched recommendations.

\subsubsection{Embedding-space visualization (UMAP)}
\label{sssec:umap}

We visualize the label, predictions, recent history, and context tokens in UMAP space. For clarity, we 1) plot user history with emphasis on more recent items, 2) jitter context sequence points to avoid occlusion and 3) zoom in to focus on the neighborhood around the predictions and ground truth point, omitting some outlying context and user points.

\begin{figure}[t]
    \centering
    \includegraphics[width=\linewidth]{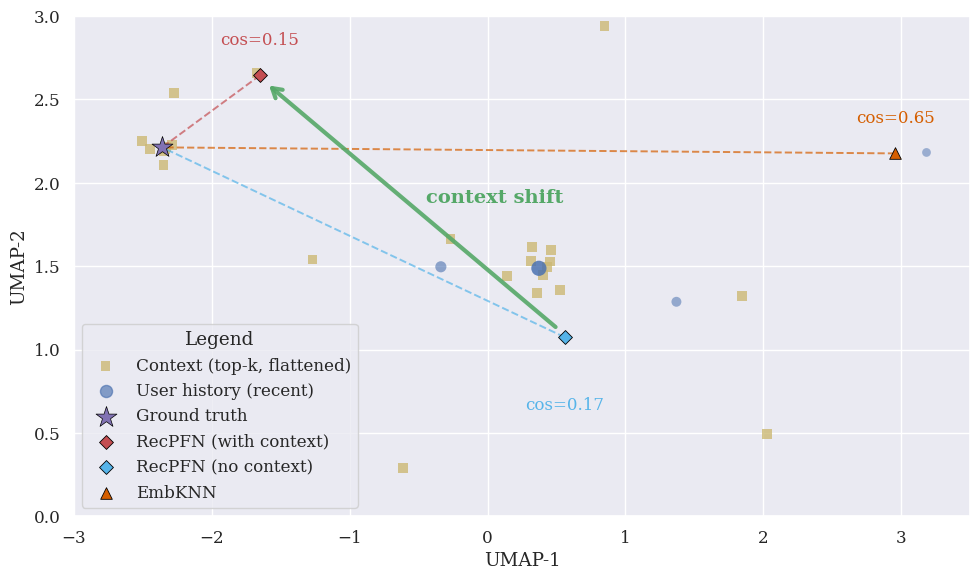}
    \caption{UMAP projection for the Amazon Movies case in Table~\ref{tab:qual_case}. Dense, relevant context near the label (star) pulls RecPFN’s prediction (blue diamond) toward the ground truth, as indicated by the arrow from the no-context prediction. Scattered context is ignored. Axes: UMAP-1 and UMAP-2.}
    \label{fig:umap_case}
\end{figure}

In this case, multiple context sequences encode the transition \emph{In the Arms of Angels} $\rightarrow$ \emph{A Pioneer Miracle}. This creates a a dense cluster of context points aligns with the ground-truth label. We see that this steers RecPFN toward the ground truth label (green arrow), while RecPFN correctly ignores scattered context points elsewhere. Without context, the prediction drifts to semantically plausible but non-specific regions; EmbKNN similarly favors broadly popular titles.


\subsection{Prior Hyperparameter Sensitivity}
\label{ssec:sensitivity}

A natural question is whether RecPFN's performance is brittle to the choice of synthetic prior: does it degrade sharply when evaluated on environments whose hyperparameters fall outside the training distribution? To assess this, we fix the trained model and sweep each SDG hyperparameter across a range spanning both inside and outside its training values, generating 50 independent synthetic environments per setting and reporting MRR@10.

\begin{figure}[t]
    \centering
    \includegraphics[width=\linewidth]{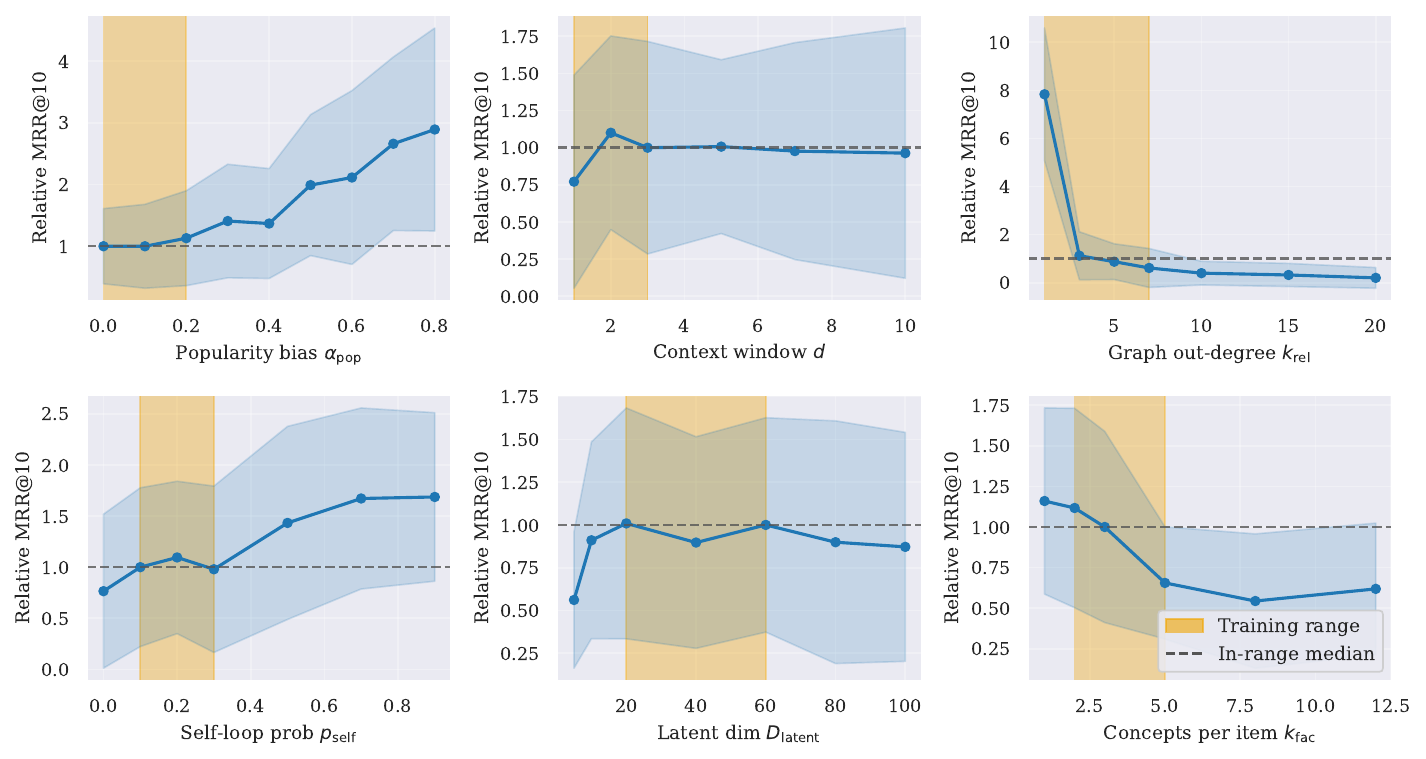}
    \caption{MRR@10 relative to the in-range median as each SDG hyperparameter is swept inside and outside its training range (shaded). Results are averaged over 50 synthetic environments per value; shaded bands show $\pm 1$ std. RG = random-graph prior; LF = latent-factor prior.}
    \label{fig:sensitivity}
\end{figure}

Figure~\ref{fig:sensitivity} summarises the results. Performance is stable across all parameters within the training range. Outside it, degradation is generally gradual. Two parameters deserve particular mention. For popularity bias ($\alpha_{\rm pop}$), out-of-range values actually \emph{increase} MRR@10: a high popularity bias concentrates sequences onto a small set of popular items, making next-item prediction trivially easier — this reflects a ceiling effect in the synthetic environments rather than a genuine robustness gain. For graph out-degree ($k_{\rm rel}$), the sharpest drop occurs at large values, where two effects compound: the transition structure becomes harder to infer from limited context, and the next-item distribution is inherently more diffuse, reducing the maximum achievable MRR@10 regardless of model quality. Overall, the results confirm that the prior provides adequate coverage for the evaluation domains, and that degradation outside it is principled rather than catastrophic.

\section{Conclusion}
\label{sec:conclusion}

We presented RecPFN, a prior-fitted, in-context learner for sequential recommendation that performs next-item prediction in a single forward pass while adapting to a new domain purely through a small set of support sequences. By pre-training on a broad synthetic prior defined by structural causal models, RecPFN amortizes Bayesian-style inference over diverse environments, avoiding domain-specific fine-tuning and large-scale real-data pre-training. Across eight public benchmarks, RecPFN achieves state-of-the-art zero-shot results and remains competitive with supervised methods in low-compute and low-data regimes, despite relying only on concise in-context adaptation at inference.

Our analysis highlights several design choices that enable these gains. Alternating attention blocks balance copying-style retrieval and feature extraction; hard-alibi masking focuses computation on recent context; and a two-stage curriculum over synthetic priors stabilizes training while capturing factor-driven dynamics. Ablations further show that replacing the synthetic prior with real interactions reduces transfer, underscoring the importance of training for generalizable inference rather than memorization.

Limitations point to clear avenues for improvement. Performance degrades when contexts become large or poorly selected, suggesting hierarchical or locality-aware routing, improved retrieval, and longer-context pre-training. Embedding quality also matters, motivating stronger or multimodal encoders and tighter alignment between training and deployment representations. Future work includes (i) richer synthetic priors with long-range and hierarchical motifs and slate/session structure; (ii) structured retrieval with jointly trained selectors to preserve signal-to-noise at larger contexts; (iii) longer-context pre-training and ICL scaling laws; (iv) stronger or multimodal encoders with contrastive alignment; and (v) analyses of amortized posterior quality, calibration, and failure modes under catalog churn and domain shift, alongside real-world A/B tests.



\appendix

\section{Synthetic data generation hyperparameters \label{apdx: sdg}}
This appendix details the sampling ranges of the environment hyperparameters for synthetic data pre-training.

\paragraph{Generic hyperparameters.}
We sample the popularity mixture $\alpha_{\text{pop}} \sim \mathcal{U}(0, 0.5)$, the context window size $d \in \{1,2,3\}$, and set recency weights $w_t=\omega^t$ with $\omega \sim \mathcal{U}(0.5, 0.8)$. The catalog size is $N{=}500$ in stage~1 and $N{=}1000$ in stage~2.

\paragraph{Random-graph prior.}
We sample $k_{\text{rel}} \in \{1,3,5,7\}$ and $p_{\text{self}} \sim \mathcal{U}(0.1, 0.3)$, and construct $\mathbf{T}$ as in Section~\ref{ssec:transition_priors}.

\paragraph{Latent-factor prior.}
We sample $D_{\text{latent}} \in \{20,40,60\}$, $k_{\text{fac}} \in \{2,3,5\}$, and $p_{\text{self}} \sim \mathcal{U}(0.1, 0.3)$. We then build $\mathbf{F}$, $\mathbf{L}_{\text{embed}}$, and $\mathbf{T}_{\text{latent}}$ (Section~\ref{ssec:transition_priors}), and project to item-level $\mathbf{E}=\mathbf{F}\mathbf{L}_{\text{embed}}$ and $\mathbf{T}=\mathbf{F}\mathbf{T}_{\text{latent}}\mathbf{F}^\top$ (with diagonal set by $p_{\text{self}}$).

\bibliographystyle{ACM-Reference-Format}
\bibliography{main}

\end{document}